\documentclass[10pt,journal]{IEEEtran}
\pdfoutput=1
\ifCLASSOPTIONcompsoc
  \usepackage[nocompress]{cite}
\else
  \usepackage{cite}
\fi
\ifCLASSINFOpdf
\else
\fi
\usepackage{times}
\usepackage{epsfig}
\usepackage{graphicx}
\usepackage{amsmath}
\usepackage{amssymb}
\usepackage{subfigure}
\usepackage{float}

\begin{document}
\title{Low-light Image Enhancement Algorithm Based on Retinex and Generative Adversarial Network}

\author{Shi~Yangming,~\IEEEmembership{Student Member,~IEEE,}
        Wu~Xiaopo,~\IEEEmembership{Member,~IEEE,}
        and~Zhu~Ming,~\IEEEmembership{Member,~IEEE}
\thanks{Manuscript received June ***, 2019; This work was supported by the Anhui Provincial Natural Science Foundation under Grant 1908085QF254.}
\IEEEcompsocitemizethanks{\IEEEcompsocthanksitem Shi~Yangming and Zhu~ming are with the Automation Department of University of Science and Technology of China, Hefei, Anhui, China, 230026.\protect\\
E-mail: ymshi@mail.ustc.edu.cn, mzhu@ustc.edu.cn
\IEEEcompsocthanksitem Wu~Xiaopo is with the Department of Electronic Engineering and Information Science  of University of Science and Technology of China, Hefei, Anhui, China, 230026. E-mail: wuxiaopo@ustc.edu.cn}}

\IEEEtitleabstractindextext{%

\begin{abstract}
Low-light image enhancement is generally regarded as a challenging task in image processing, especially for the complex visual tasks at night or weakly illuminated. In order to reduce the blurs or noises on the low-light images, a large number of papers have contributed to applying different technologies. Regretfully, most of them had served little purposes in coping with the extremely poor illumination parts of images or test in practice. In this work, the authors propose a novel approach for processing low-light images based on the Retinex theory and generative adversarial network~(GAN), which is composed of the decomposition part for splitting the image into illumination image and reflected image, and the enhancement part for generating high-quality image. Such a discriminative network is expected to make the generated image clearer. Couples of experiments have been implemented under the circumstance of different lighting strength on the basis of Converted See-In-the-Dark~(CSID) datasets, and the satisfactory results have been achieved with exceeding expectation that much encourages the authors. In a word, the proposed GAN-based network and employed Retinex theory in this work have proven to be effective in dealing with the low-light image enhancement problems, which will benefit the image processing with no doubt.
\end{abstract}
}
\maketitle
\IEEEdisplaynontitleabstractindextext
\IEEEpeerreviewmaketitle

\begin{figure}[htbp]
\begin{center}
  \subfigure[Input]{
    \begin{minipage}[b]{0.21\linewidth}
        \includegraphics[width=\linewidth]{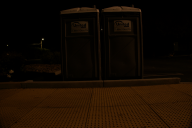}\vspace{6pt}
        \includegraphics[width=\linewidth]{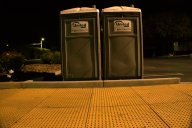}
    \end{minipage}
  }
  \subfigure[LIME]{
    \begin{minipage}[b]{0.21\linewidth}
      \includegraphics[width=\linewidth]{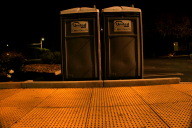}\vspace{6pt}
      \includegraphics[width=\linewidth]{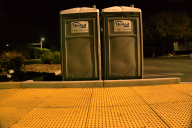}
    \end{minipage}
  }
  \subfigure[Input]{
    \begin{minipage}[b]{0.21\linewidth}
      \includegraphics[width=\linewidth]{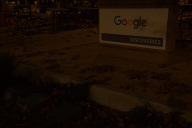}\vspace{6pt}
      \includegraphics[width=\linewidth]{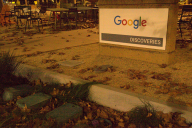}
    \end{minipage}
  }
  \subfigure[LightenNet]{
    \begin{minipage}[b]{0.21\linewidth}
      \includegraphics[width=\linewidth]{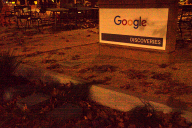}\vspace{6pt}
      \includegraphics[width=\linewidth]{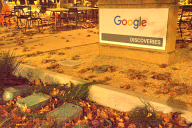}
    \end{minipage}
  }
\end{center}
\caption{Experimental results by LIME and LightenNet. The input images of first line are with brightness levels of 0.1, and the second line are with brightness levels of 1.}
\label{fig:failure}
\end{figure}

\ifCLASSOPTIONcompsoc
\IEEEraisesectionheading{\section{Introduction}\label{sec:introduction}}
\else
\section{Introduction}
\label{sec:introduction}
\fi
For the past several years, deep convolution neural networks (DCNNs) have found their extensive applications in image processing such as image classification~\cite{he2016deep,Huang_2017_CVPR,krizhevsky2012imagenet,simonyan2014very,szegedy2015going}, object detection~\cite{he2017mask,lin2017focal,liu2016ssd,redmon2018yolov3,ren2015faster,zhao2018m2det,rezatofighi2019generalized}, image segmentation~\cite{Iglovikov_2018_CVPR_Workshops, chen2017deeplab,zhao2017pyramid,zhou2017scene,he2017mask} and object tracking~\cite{bertinetto2016fully,chen2018deeplab,he2018twofold,li2018structure,luo2017end,ristani2018features}, and so forth. And thanks to the development of DCNNs, it has sparked remarkable activity in specific image analysis. While, in spite of tremendous progress brought by DCNNs, one unified theory and general solution of low light image processing had eluded the most researchers, especially for the cases of poor illumination condition and camera shaking in the real world. There is still a substantial number of scientific papers contributing to the examinations on low-light enhancement task~\cite{fu2016fusion,fu2016weighted,guo2017lime,jobson1997multiscale,li2018lightennet,lore2017llnet,wang2016fusion,wei2018deep,shen2017msr} in the latest years. Interesting though those works were, these approaches can achieve good results under certain conditions but still have some limitations. The greatest challenge is that there was no suitable dataset for training and testing. Those researchers usually experimented on artificial low-light datasets needing darken-process and denoise-process on the original images, which resulted in an embarrassment that images from real environment can’t be processed effectively. When dealing with the image of relatively poor lightness and be with ambient noise, these approaches can't get satisfactory results. Fig.~\ref{fig:failure} shows part of the results by some of those methods. According to the images generated by the LIME~\cite{guo2017lime} method, one could find that the lower brightness images can't be restored well in detail, such as the top right-hand corner distant dark trees. The results by the LightenNet~\cite{li2018lightennet} method show that the method is not stable enough, the generated images are overexposed and blurred in high illumination level and full of noise for the case of low-light illumination.

To avoid the aforementioned dilemma, the authors propose a novel method based on cross-domain algorithms. It is assumed that images from low-light environment belong to dark domain in a same distribution while long exposure or normal images belong to another domain which satisfy another distribution. We can use cross-domain approach to solve this low-light enhancement problem. It is the generative adversary networks (GANs)~\cite{goodfellow2014generative} that make the cross-domain image-to-image be so attractive and considerable numbers of approaches based on GANs are generally employed to deal with domain-transferring issues.  Years of research in computer vision, image processing, computational photography, and graphics have produced powerful translation systems. These approaches can be divided into two categories, one is the supervised algorithm~\cite{isola2017image} and the other is the unsupervised one~\cite{liu2017unsupervised,yi2017dualgan,zhu2017unpaired}. In the former method, paired inputs are needed to obtain higher performance. While, in the latter method, paired inputs are not necessary and one could readily collect plenty of the unsupervised datasets. By comparison, we prefer to adopt the supervised algorithm not only for the brought higher performance, but also due to the fact that the unsupervised algorithm fail to get precise results in detail when dealing with complex tasks.

The supervised algorithm we suggest is actually utilizing a hybrid architecture based on generative adversarial network and Retinex theory~\cite{land1971lightness}. In order to employ effectively the Retinex theory in generative adversary networks, the authors had tried three different strategies for applying the Retinex to CNNs and developing the regularization loss to avoid the local optimal solution. Fortunately, the experimental results indeed help to verify the available algorithm. Meanwhile, to maximize the performance of the model, the \textbf{LO}w-\textbf{L}ight (LOL) dataset produced by Wei \textit{et al.}~\cite{wei2018deep} and the \textbf{C}onverted \textbf{S}ee-\textbf{I}n-the-\textbf{D}ark (CSID) dataset (originating from the raw image of See-In-the-Dark) were introduced to implement the experiment. The satisfactory results demonstrate that Retinex-GAN could bring about considerable improvement as expected.

In this paper’s remaining, Section II describes the related work of GAN and image enhancement to this research in retrospect. And in Section III, the proposed network and the dataset production approach are presented detailedly, followed by the experimental results and discussions in Section IV.

\section{Background and Previous work}
\subsection{Low-light Enhancement and Image Denoising}

The past decades have witnessed the rapid development of low-light image enhancement from unpopular domain to hot topics, owing to the booming techniques of deep learning. There exist lots of approaches proposed to improve the quality of low-light images. To sum up, those approaches could be generally divided into three categories, namely the pre-processing, the post-processing, and using both of them. The pre-processing methods are mainly to use algorithms or physical setup, such as flash photography techniques. Other well-known methods include histogram equalization~(HE)~\cite{land1971lightness}, Contrast-Limiting Adaptive Histogram Equalization~(CLAHE)~\cite{reza2004realization}, Unsharp Masking~(UM)~\cite{solomon2011fundamentals}, Multiscale Retinex~(MSR)~\cite{jobson1997multiscale}, Msr-net: Low-light image enhancement using deep convolutional network~(Msr-net)~\cite{shen2017msr}, a weighted variational model for simultaneous reflectance and illumination estimation~(AWVM)~\cite{fu2016weighted}, low-light image enhancement~(LIME)~\cite{guo2017lime}, the low-light net~(LLNET)~\cite{lore2017llnet}, LightenNet~\cite{li2018lightennet} and \textit{etc.}.

HE is a popular algorithm for image enhancement which adjusts the intensity of image for better quality. CLAHE is a method developed based on adaptive histogram equalization~(AHE) which transforms the intensity of the pixel into the display range proportional to the pixel intensity’s rank in the local intensity histogram, and will reduce the effect of edge shadowing. UM is a method that used for sharpening image quality by blurring and then adding some differences to the original image. BM3D is an algorithm for noise removal by utilizing Wiener filter as a collaborative form used to filter dimensional patches block by clustering similar blocks from 2D to 3D array of data and afterward denoising the gathered fixes mutually. Then, the denoised patches are connected back to the first pictures by a voting instrument which expels noise from the considered area.

Besides, many deep-learning based algorithms are proposed to deal with those issues and found themselves with overwhelming advantages. Lore~\textit{et al.} proposed LLNet~\cite{lore2017llnet} to learn underlying signal features in low-light images by using deep AutoEncoders. Fu~\textit{et al.}~\cite{fu2016weighted} proposed a weighted minimization algorithm for estimating reflectance and illumination from an image. Guo~\textit{et al.}~\cite{guo2017lime} developed a structure-aware smoothing model to improve the illumination consistency of images. Lore~\textit{et al.}~\cite{lore2017llnet} proposed a deep AutoEncoder approach to learn features from low-light images and then enhance those images. Li~\textit{et al.}~\cite{li2018lightennet} proposed the LightenNet which learns a mapping between weakly illuminated image and the corresponding illumination map to obtain the enhanced image. Meanwhile, Wei~\textit{et al.}~\cite{wei2018deep} proposed a deep Retinex decomposition method which can learn to decompose the observed image into reflectance and illumination in a data-driven way without decomposing further the image of ground truth. The authors tried to improve the algorithm to make the process of image-brightening and de-noising more effectively. 

\subsection{Generative Adversarial Networks~(GAN)}

Recently, generative adversarial nets~(GAN)~\cite{goodfellow2014generative} have been attracting most of the attentions in image-to-image translation. The original GAN includes two networks which are built to play a one-sum games. The generative network is trained for generating realistic synthetic samples from a noise distribution to cheat the discriminative network. The discriminative network aims to distinguish true and generated fake samples. In addition to random sample from a noise distribution, various form data can also be used as input to the generator.

Image-to-image translation algorithms transfer an input image from one domain to corresponding image in another domain. Isola~\textit{et al.}~\cite{isola2017image} first proposed supervised GAN which used U-net as the generative network and make good results in domain transfer. Zhu~\textit{et al.}~\cite{zhu2017unpaired} provided an unsupervised algorithm which mapped images from one domain to another domain and then mapped to the original domain, and finally used cycle-consistent loss to reduce the difference. Simultaneously, Yi~\textit{et al.}~\cite{yi2017dualgan}, Liu~\textit{et al.}~\cite{liu2017unsupervised} and Kim~\textit{et al.}~\cite{kim2017learning} had put emphasis on the domain transfer with unsupervised manners.

What the authors trying to examine was inspired by those image-to-image translation approaches. We assume that images from dark environments with noise which were meeting a distribution and coming from one specific domain, and the high-quality images were from another domain. The recovery step is an image-to-image translation process. To the best of our knowledge, applying generative adversary networks to the low-light image enhancement and image denoising still remains an opening area, the authors are the first to propose such a novel technique to cope with the low light image processing and to obtain the results with exceeding expectations.

\section{Proposed Approach}
\subsection{Network Structure}
  \begin{figure*}[htbp]
	\centering
	\includegraphics[width=\linewidth]{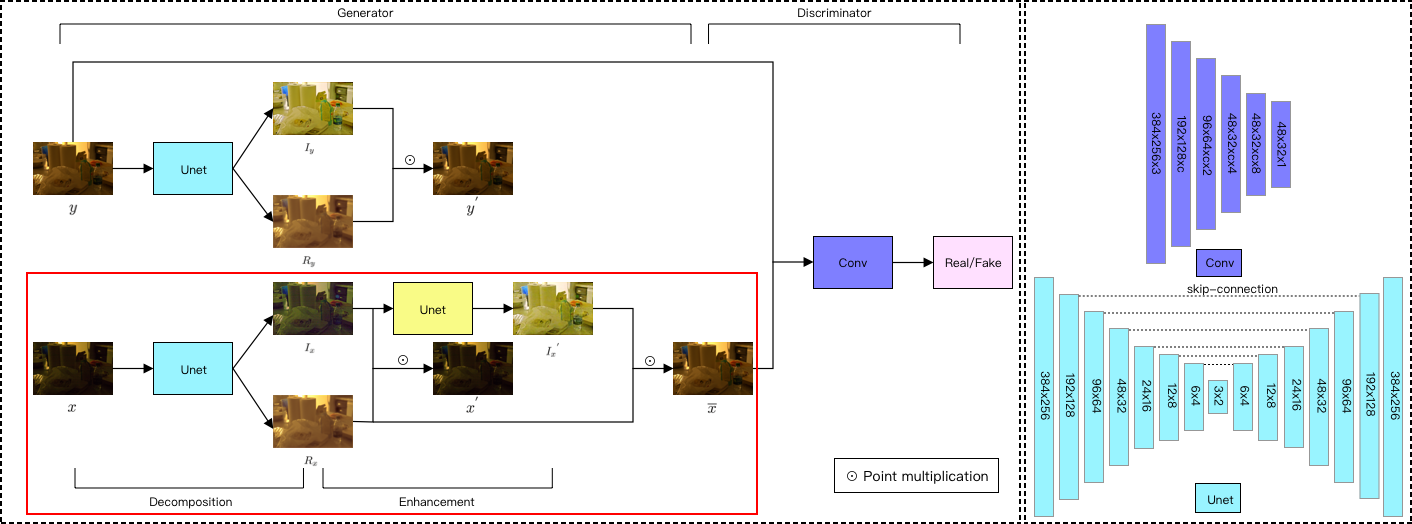}

	\caption{Retinex-GAN: $x$ is low-light image and $y$ is the corresponding ground truth image. The generative network is composed of two parallel Unets to split $x$ and $y$ into reflected image when data training commences, which is also termed as the decomposition process. The following enhancement process in the generative network is responsible for generating the reflected images for $x$ by the yellow Unet and then a new image is formed by combining the reflected part and illumination part. At last, the discriminative network, actually a normal convolutional neural network, is to distinguish $\overline{x}$ and $y$. When testing, only the area encircled by red line rectangular works.}
	\label{fig:network}
\end{figure*}

\begin{figure}
  \centering
  
    \includegraphics[width=0.7\linewidth]{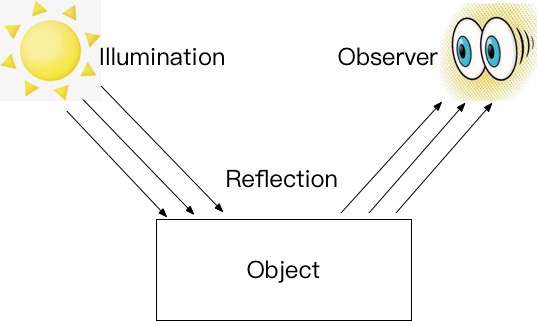}
    \caption{Illustration of Retinex theory. The image observed can be decomposed into the brightness matrix produced by the light source and the reflectivity matrix of the object.}
    \label{fig:retinex}
  \end{figure}

Our approach benefits from the Retinex theory by~E.H.Land~\cite{land1971lightness}. It is well known in Retinex theory that the color of an object is determined by its ability to reflect long-wave~(\textbf{red}), medium-wave~(\textbf{green}) and short-wave~(\textbf{blue}) light rather than the absolute value of the reflected light intensity. The color of an object is not affected by the illumination heterogeneity and has consistency, that is, Retinex theory is based on the color consistency~(color\ constancy). As shown in Fig.~\ref{fig:retinex}, the image S is obtained by reflecting the incident light~$L$ from the surface of the object. The reflectivity~$R$ is determined by the object itself. It’s assumed in Retinex theory that the original image~$S$ consist of the product of the illumination image~$I$ and the reflected image~$R$, which can be expressed as the following form:
\begin{equation}
    S(i,j) = R(i,j)I(i,j)
  \end{equation}
  where $(i,j)$ represent the pixel location.

We apply the Retinex theory to our neural network. As shown in Fig.~\ref{fig:network}, the network is composed of one generative network and one discriminative network. The generative network~(\textbf{G}) which looks like a letter \textbf{W} includes decomposition part and enhancement part. The decomposition part aims to split the original image into illumination image~$I_x$ and reflected image~$R_x$ while the enhancement part tries to enhance the brightness of the image. Finally, the reflected image~$R_x$ and new illumination image~${I_x}^{'}$ do dot product operation and output a new normal image. Meanwhile, the discriminative network~(\textbf{D}) make the generated image look more realistic which have ability on distinguishing noised image from real high-quality images. Let $x \in \mathcal{X}$ be low-light image, $y \in \mathcal{Y}$ be high-quality image. The whole network aims to recover images in domain $\mathcal{Y}$ from corresponding images in domain $\mathcal{X}$. We simply call it Retinex-GAN for a convenience.

\subsection{Regularization Loss}
\label{ref:regular}
  We assume that RGB channels of images are exposed to the same level of light, thus we use the Unet to split the image $S$ into reflected image $R$ with three channels and illumination image $I$ with one channel. It has proved that the assumption fails to maximize the network performance while illumination image $I$ and reflected image $R$ can't reconstruct the original image $S$ effectively~\cite{land1971lightness}. Then we try the second strategy. We assume that the illumination image is also three channels. On the basis of this assumption, there is a serious problem that the network quickly falls into a local optimal solution. Let's give a brief analysis of this problem from a mathematical point of view. Given two matrix $S_1,S_2$, if we want optimize $min|R_1-R_2|$ to satisfy the following equation for all points $i,j$:
  \begin{equation}
    \left\{
  \begin{array}{rcl}
    S_1(i,j) = R_1(i,j)I_1(i,j) \\
    S_2(i,j) = R_2(i,j)I_2(i,j)
  \end{array} \right.
  \end{equation}
  Obviously, the best two solutions are as below:
  \begin{equation}
    \left\{
  \begin{array}{rcl}
    R_1(i,j) = 1 \\
    R_2(i,j) = 1 \\
    I_1(i,j) = S_1(i,j) \\
    I_2(i,j) = S_2(i,j)
  \end{array} \right.
  \end{equation}
  and
  \begin{equation}
    \left\{
  \begin{array}{rcl}
    R_1(i,j) = -1 \\
    R_2(i,j) = -1 \\
    I_1(i,j) = -S_1(i,j) \\
    I_2(i,j) = -S_2(i,j)
  \end{array} \right.
  \end{equation}
  \\According to the above inference, after many iterations, the value of reflected image will become all 1 or -1 and the illumination image will be same as original or the reversed image. This means that the decomposition part does useless work. To solve this problem, we propose a regularization loss $\mathcal{L}_{reg}$ which can prevent the RGB values of generated illumination image from approaching 1 or -1 to avoid that the network falls into a local optimal solution.
  \begin{equation}
    \mathcal{L}_{reg} = \frac{1}{mn}\sum_{i=0}^{m-1}\sum_{j=0}^{n-1}\frac{1}{C-f(R(i,j))} \qquad C\ge1
  \end{equation}
  If $R(i,j)$ approaching to $C\approx1\ or\ C\approx-1$, $\mathcal{L}_{reg}$ becomes large which makes $R(i,j)$ far away from 1 or-1. The regularization loss is quite useful during training and testing.

\subsection{Multitask Loss}
  Formally, during training, we define a multi-task loss as
  \begin{equation}
    \mathcal{L} = \lambda_{rec}\mathcal{L}_{rec} + \lambda_{dec}\mathcal{L}_{dec} + \lambda_{com}\mathcal{L}_{com} + \lambda_{cGAN}\mathcal{L}_{cGAN}
  \end{equation}
  where $\lambda_{rec}$, $\lambda_{dec}$, $\lambda_{com}$ and $\lambda_{cGAN}$ (short for condition GAN) are respectively the loss weightings for each loss term.
  We take part of the loss of Pix2pix-GAN which includes the $\mathcal{L}_{cGAN}$ loss and the $\mathcal{L}_1$ loss while the $\mathcal{L}_1$ loss is replaced by $smooth \mathcal{L}_1$ loss. The original $cGAN$ loss can be described as:
  \begin{equation}
    \mathcal{L}_{cGAN}(G,D) = \mathbb{E}_{x,y}[\log D(x,y)]+\mathbb{E}_{x}[\log(1-D(x,G(x)))]
  \end{equation}
  The $smooth \mathcal{L}_1$ loss is defined as:
  \begin{equation}
    \mathcal{L}_{\mathcal{L}_{1}}(x,y) = smooth_{\mathcal{L}_1}(x,y)
  \end{equation}
  in which
  \begin{equation}
    smooth_{\mathcal{L}_1}(x) = \left\{
    \begin{array}{rcl}
      0.5x^2 & &if\quad x<1\\
      x - 0.5 & &otherwise,
    \end{array} \right.
  \end{equation}
  The reconstruction loss $\mathcal{L}_{rec}$ ensures that the image divide into illumination part and reflected part then can be restored which is defined as:
  \begin{equation}
    \mathcal{L}_{rec} = \mathcal{L}_{rec\_x} + \mathcal{L}_{rec\_y} + \mathcal{L}_{reg}
  \end{equation}
  where
  \begin{equation}
    \mathcal{L}_{rec\_x} = \mathcal{L}_{\mathcal{L}_{1}}(x , R_x \cdot I_x)
  \end{equation}
  \begin{equation}
    \mathcal{L}_{rec\_y} = \mathcal{L}_{\mathcal{L}_{1}}(y , R_y \cdot I_y)
  \end{equation}
  The decomposition loss is defined as:
  \begin{equation}
    \mathcal{L}_{dec} = \mathcal{L}_{\mathcal{L}_{1}}(I_x , I_y)
  \end{equation}
  The decomposition loss makes the image in different brightness is decomposed to the same illumination images. And finally the enhancement loss optimize the $\mathcal{L}_{1}$ distance of composite image and target image which can be described as:
  \begin{equation}
    \mathcal{L}_{enh} = \mathcal{L}_{\mathcal{L}_{1}}(y, \overline{x}) = \mathcal{L}_{\mathcal{L}_{1}}(y, R_x \cdot {I_x}^{'})
  \end{equation}
  where $R^{'}$ is the enhanced reflected image.

  In order to obtain the image details, we use the better SSIM-MS loss $\mathcal{L}_{ssim-ms}$ which are proposed by Zhao~\textit{et al.}~\cite{zhao2017loss}. The SSIM-MS loss is a multi-scale version of SSIM loss which comes from the \textbf{S}tructural \textbf{SIM}ilarity index~(SSIM). Means and standard deviations are computed with a Gaussian filter with standard deviation $\sigma_G$,$G_{\sigma_G}$. SSIM for pixel p is defined as:
  \begin{equation}
    SSIM(i,j) = \frac{(2\mu_i\mu_j + c_1)(2\sigma_{ij} + c_2)}{(\mu_i^2 + \mu_j^2 + c_1)(\sigma_i^2 + \sigma_j^2 + c_2)}
    \label{equ:ssim}
  \end{equation}
  where $\mu_i,\mu_j$ is the average of $i,j$, $\sigma_i^2,\sigma_j^2$ is the variance of $i,j$, $\sigma_{ij}$ is the convariance of $i$ and $j$, $c_1={(k_1L)}^2,c_2={k_2L}^2$ are two variables to stabilize the division with weak denominator, $L$ is the dynamic range of the pixel-values, $k_1=0.01$ and $k_2=0.03$ by default.
  The loss function for SSIM can be then written as:
  \begin{equation}
    \mathcal{L}_{ssim}(p) = \frac{1}{mn}\sum_{i=0}^{m-1}\sum_{j=0}^{n-1}(1-SSIM(i,j)) =l(p)cs(p)
  \end{equation}
  while SSIM Loss are influenced by the parameters $\sigma_G$, Zhao~\textit{et al.} use the $MS\_SSIM$ rather than fine-tuning the $\sigma_G$, Given a dyadic pyramid of M levels, $MS\_SSIM$ is defined as:
  \begin{equation}
    MS\_SSIM(p) = l_M^{\alpha}(p)\prod_{j=1}^M
    cs_j^{\beta_j}(p)
  \end{equation}
  The multiscale SSIM loss for patch p is defined as:
  \begin{equation}
    \mathcal{L}_{ssim\_ms} = 1 - {MS\_SSIM}(\tilde{p})
  \end{equation}
  We combine the $\mathcal{L}_{enh}$ with $\mathcal{L}_{ssim\_ms}$ and take the strategies in~\cite{zhao2017loss}.
  \begin{equation}
    \mathcal{L}_{com} = {\alpha}\mathcal{L}_{enh}+(1-\alpha)\mathcal{L}_{ssim\_ms}
  \end{equation}
  where $\alpha$ is set to 0.84.

\section{Experimental Results and Discussions}
  In order to verify the framework mentioned above, large numbers of CSID dataset~\cite{chen2018learning} and the LOL dataset~\cite{wei2018deep} were chosen for the experiment. All the models are implemented with the Tensorflow framework on 1080Ti GPUs. Training details and structure models from Pix2pix-GAN will help to build our Retinex-GAN. The resolution of the images for all the experiments is set to 384 x 256. The generative network is composed of two similar Unets~\cite{ronneberger2015u} from generated network of Pix2pix-GAN~\cite{isola2017image} and all Unets adopt skip-connection strategies. The discriminator network is a ${46}\times{46}$ PatchGAN that is used to distinguish a ${46}\times{46}$ image patch whether it is real or fake. We use Adam optimizer with the proposed optimization settings in~\cite{kingma2014adam} with $[\beta_1, \beta_2, \epsilon] = [0.5, 0.999, 10^{-8}]$. The batch size is set to 16. The initial learning rate is 0.0002 for all the trained network, which would be decreased manually when the training loss converges. For the loss weighting in our final loss function, we empirically find that the combination $[\lambda_{rec}, \lambda_{dec}, \lambda_{com}, \lambda_{cGAN}] = [1, 1, 10, 1]$ results in a stable training.

\subsection{Converted See In the Dark Dataset}
\begin{figure}[htbp]
    \centering
      \includegraphics[width=0.22\linewidth]{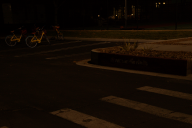}\vspace{6pt}
      \includegraphics[width=0.22\linewidth]{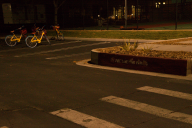}
      \includegraphics[width=0.22\linewidth]{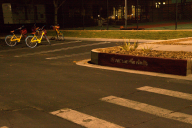}
      \includegraphics[width=0.22\linewidth]{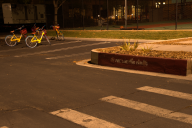}
      \includegraphics[width=0.22\linewidth]{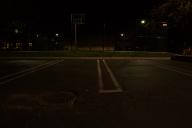}\vspace{6pt}
      \includegraphics[width=0.22\linewidth]{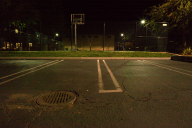}
      \includegraphics[width=0.22\linewidth]{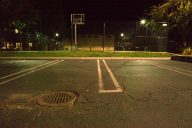}
      \includegraphics[width=0.22\linewidth]{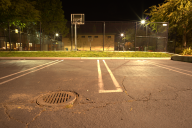}
      \includegraphics[width=0.22\linewidth]{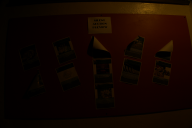}
      \includegraphics[width=0.22\linewidth]{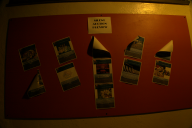}
      \includegraphics[width=0.22\linewidth]{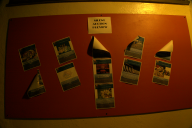}
      \includegraphics[width=0.22\linewidth]{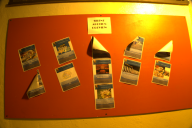}
    \caption{Examples in the CSID datasets: from left to right, the brightness levels are: 1(0.1), 3(0.5), 5(0.9), and the ground truth.}
    \label{fig:datasetsimg}
  \end{figure}

\begin{figure*}
    \centering
    \subfigure[Input]{
    \begin{minipage}[b]{0.124\linewidth}
      \includegraphics[width=\linewidth]{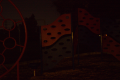}\vspace{6pt}
      \includegraphics[width=\linewidth]{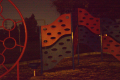}\vspace{6pt}
      \includegraphics[width=\linewidth]{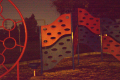}\vspace{6pt}
      \includegraphics[width=\linewidth]{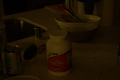}\vspace{6pt}
      \includegraphics[width=\linewidth]{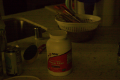}\vspace{6pt}
      \includegraphics[width=\linewidth]{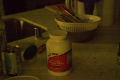}
    \end{minipage}
    }
    \subfigure[CLAHE~\cite{reza2004realization}]{
    \begin{minipage}[b]{0.124\linewidth}
      \includegraphics[width=\linewidth]{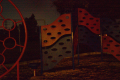}\vspace{6pt}
      \includegraphics[width=\linewidth]{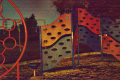}\vspace{6pt}
      \includegraphics[width=\linewidth]{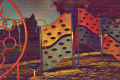}\vspace{6pt}
      \includegraphics[width=\linewidth]{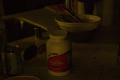}\vspace{6pt}
      \includegraphics[width=\linewidth]{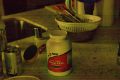}\vspace{6pt}
      \includegraphics[width=\linewidth]{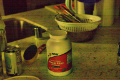}
    \end{minipage}
    }
    \subfigure[LightenNet\cite{li2018lightennet}]{
      \begin{minipage}[b]{0.124\linewidth}
        \includegraphics[width=\linewidth]{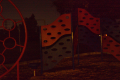}\vspace{6pt}
        \includegraphics[width=\linewidth]{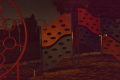}\vspace{6pt}
        \includegraphics[width=\linewidth]{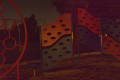}\vspace{6pt}
        \includegraphics[width=\linewidth]{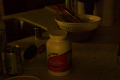}\vspace{6pt}
        \includegraphics[width=\linewidth]{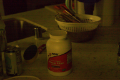}\vspace{6pt}
        \includegraphics[width=\linewidth]{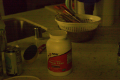}
      \end{minipage}
    }
    \subfigure[AWVM\cite{fu2016weighted}]{
      \begin{minipage}[b]{0.124\linewidth}
        \includegraphics[width=\linewidth]{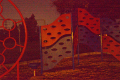}\vspace{6pt}
        \includegraphics[width=\linewidth]{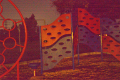}\vspace{6pt}
        \includegraphics[width=\linewidth]{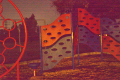}\vspace{6pt}
        \includegraphics[width=\linewidth]{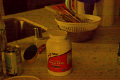}\vspace{6pt}
        \includegraphics[width=\linewidth]{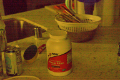}\vspace{6pt}
        \includegraphics[width=\linewidth]{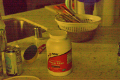}
      \end{minipage}
    }
    \subfigure[LIME~\cite{guo2017lime}]{
      \begin{minipage}[b]{0.124\linewidth}
        \includegraphics[width=1\linewidth]{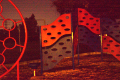}\vspace{6pt}
        \includegraphics[width=1\linewidth]{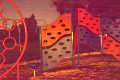}\vspace{6pt}
        \includegraphics[width=1\linewidth]{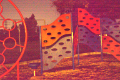}\vspace{6pt}
        \includegraphics[width=1\linewidth]{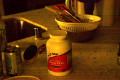}\vspace{6pt}
        \includegraphics[width=1\linewidth]{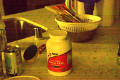}\vspace{6pt}
        \includegraphics[width=1\linewidth]{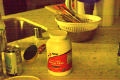}
      \end{minipage}
    }
    \subfigure[Ours]{
      \begin{minipage}[b]{0.124\linewidth}
        \includegraphics[width=\linewidth]{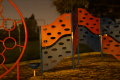}\vspace{6pt}
        \includegraphics[width=\linewidth]{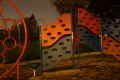}\vspace{6pt}
        \includegraphics[width=\linewidth]{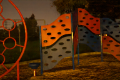}\vspace{6pt}
        \includegraphics[width=\linewidth]{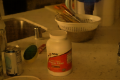}\vspace{6pt}
        \includegraphics[width=\linewidth]{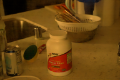}\vspace{6pt}
        \includegraphics[width=\linewidth]{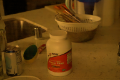}
      \end{minipage}
    }
    \subfigure[Ground truth]{
      \begin{minipage}[b]{0.124\linewidth}
        \includegraphics[width=\linewidth]{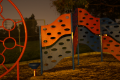}\vspace{6pt}
        \includegraphics[width=\linewidth]{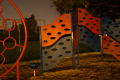}\vspace{6pt}
        \includegraphics[width=\linewidth]{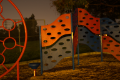}\vspace{6pt}
        \includegraphics[width=\linewidth]{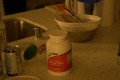}\vspace{6pt}
        \includegraphics[width=\linewidth]{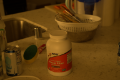}\vspace{6pt}
        \includegraphics[width=\linewidth]{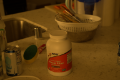}
      \end{minipage}
    }
    \caption{Visual effects comparison of experimental results on the CSID dataset. From top to bottom, the brightness levels are 0.1, 0.5 and 0.9.}
    \label{fig:resultcompare}
 \end{figure*}

 The original SID~\cite{chen2018learning} dataset is composed of 5094 raw short-exposure images, each with a corresponding long-exposure reference images and that multiple short-exposure images can correspond to the same long-exposure images. These images will be trained as the dataset for our experiments. Actually, we convert both short-exposure and long-exposure images from raw format to PNG format. The long-exposure raw images can generate high-quality PNG images by default converting free of further processing. But short-exposure raw images will show all black after converting for the lack of adequate. So we carry out some manual processing by adjusting parameter in the rawpy python package. The brightness levels are divided into 5 levels from 0.1 to 0.9 with the interval of 0.2 and the brightness value equal to 1.0 is referred as the top level. And the level 1 usually looks like the same as the long-exposure ground truth images with noise. In order to acquire the qualified input, we simply resize all converted images to ${384}\times{256}$ resolution for training and testing. On the premise of the above operation, we select 1550 pairs of images for training and 217 pairs of images for testing by removing the images with a lot of noise. Some examples of the dataset are shown in Fig.~\ref{fig:datasetsimg}.
    \\\textbf{Qualitative Analysis.} We compare our Retinex-GAN with five low-light enhancement methods including AWVM~\cite{fu2016weighted}, and CLAHE~\cite{reza2004realization}, and the state-of-the-art LIME~\cite{guo2017lime}, and LightenNet~\cite{li2018lightennet} on five different brightness level. As shown in Fig.~\ref{fig:resultcompare}, our method works best visually and the stable results are most similar to ground truth. The results show that the LightenNet method has no effect on our dataset, and even reduces the brightness of the input image. Although other methods, such as the LIME, AWVM and CLAHE, can enhance images to a certain extent owing to the fact that their experiment datasets are synthesized by reducing the value of V channel in clear illuminated images, their framework doesn't take into account the effects of noise. When testing on the real image datasets with noise by image sensors, the resulting image contains a lot of noise. Another point worth noting is that other methods will get different results when dealing with different lighting images. For instance, the CLAHE~ method try to get dim results when processing extremely poor light images.
 \\\textbf{Quantitative Analysis.} Due to the ground truth of every image is known, we can evaluate the quality of the generated image numerically. We evaluate our method with three indications. The \textbf{M}ean \textbf{S}quare \textbf{E}rror~(MSE) of generated image $A$ and ground truth image $B$ is defined by equation:
    \begin{equation}
    MSE = \frac{1}{mn}\sum_{i=0}^{m-1}\sum_{j=0}^{n-1}[I_A(i,j)-I_B(i,j)]^{2}
    \end{equation}
    where $I_A(i,j)$ and $I_B(i,j)$ are separately the pixel values of $A$ and $B$ in position (i,j).
    Then, the \textbf{P}eak \textbf{S}ignal to \textbf{N}oise \textbf{R}atio~(PSNR) is defined by equation:
    \begin{equation}
      PSNR = 10\lg{\frac{\max(I)^2}{MSE}}
    \end{equation}
    The \textbf{S}tructure \textbf{SIM}ilarity index~(SSIM) is a method for predicting quality of digital images. SSIM is used for measuring the similarity between two images which is defined by~(\ref{equ:ssim}).

\begin{figure*}[htbp]
      \centering
      \subfigure[MSE]{
        \begin{minipage}[b]{0.31\linewidth}
        \includegraphics[width=\linewidth]{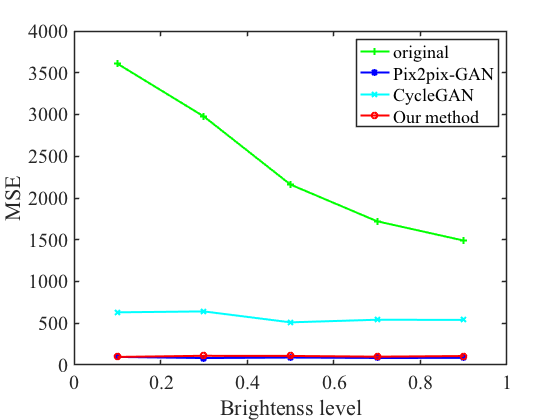}
        \end{minipage}
      }
      \subfigure[PSNR]{
        \begin{minipage}[b]{0.31\linewidth}
        \includegraphics[width=\linewidth]{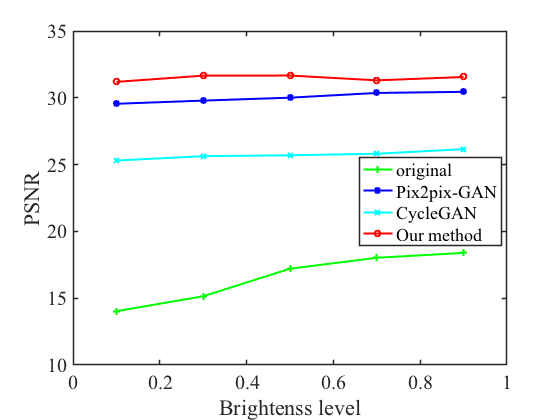}
        \end{minipage}
      }
      \subfigure[SSIM]{
        \begin{minipage}[b]{0.31\linewidth}
        \includegraphics[width=\linewidth]{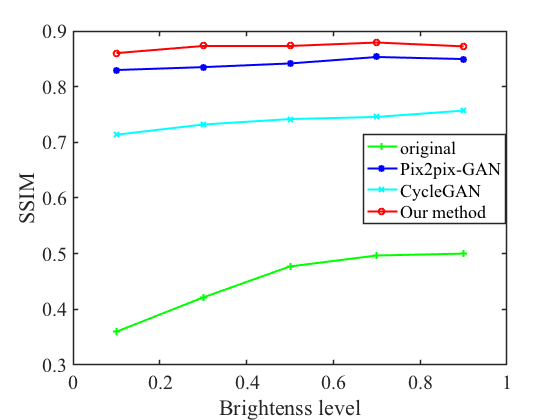}
        \end{minipage}
      }
      \caption{Comparison of numerical indicators on the CSID dataset.}
      \label{fig:mertrics}
\end{figure*}

    Because other low-light methods don't train on the CSID dataset or rely on the learning-based techniques, we only compare our method with GAN-based methods including Pix2pix-GAN~\cite{isola2017image} and CycleGAN~\cite{zhu2017unpaired} on 5 situations which the brightness levels are from 0.1 to 0.9 with the interval of 0.2. In Fig.~\ref{fig:mertrics}, thanks to the supervised learning, the result of our method is very close to ground truth as the supervised Pix2pix-GAN method while the unsupervised CycleGAN method try to approach the ground truth. 

    The MSE value of our method is slightly higher than the Pix2pix-GAN method, but our method is much better than the Pix2pix-GAN and CycleGAN methods in terms of the value of the signal-to-noise ratio (PSNR) and the value of the image similarity (SSIM). Simultaneously, the experimental results also prove to some extent the effectiveness of GAN in image enhancement.

\subsection{LOL Dataset}

\begin{figure*}[htbp]
      \centering
      \subfigure[Input]{
        \begin{minipage}[b]{0.124\linewidth}
        \includegraphics[width=\linewidth]{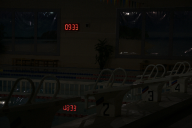}\vspace{6pt}
        \includegraphics[width=\linewidth]{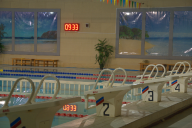}
        \end{minipage}
      }
      \subfigure[R by Retinex-Net]{
        \begin{minipage}[b]{0.124\linewidth}
        \includegraphics[width=\linewidth]{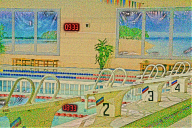}\vspace{6pt}
        \includegraphics[width=\linewidth]{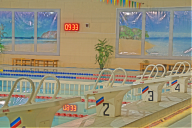}
        \end{minipage}
      }
      \subfigure[I by Retinex-Net]{
        \begin{minipage}[b]{0.124\linewidth}
        \includegraphics[width=\linewidth]{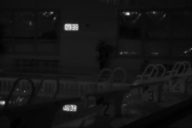}\vspace{6pt}
        \includegraphics[width=\linewidth]{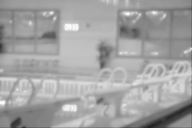}
        \end{minipage}
      }
      \subfigure[R by LIME]{
        \begin{minipage}[b]{0.124\linewidth}
        \includegraphics[width=\linewidth]{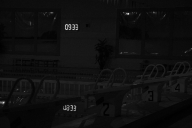}\vspace{6pt}
        \includegraphics[width=\linewidth]{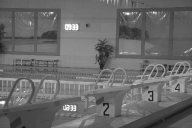}
        \end{minipage}
      }
      \subfigure[I by LIME]{
        \begin{minipage}[b]{0.124\linewidth}
        \includegraphics[width=\linewidth]{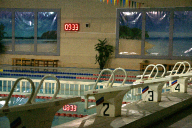}\vspace{6pt}
        \includegraphics[width=\linewidth]{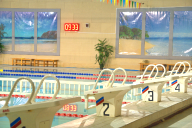}
        \end{minipage}
      }
      \subfigure[R by ours]{
        \begin{minipage}[b]{0.124\linewidth}
        \includegraphics[width=\linewidth]{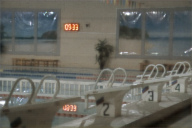}\vspace{6pt}
        \includegraphics[width=\linewidth]{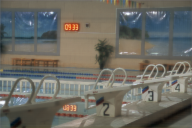}
        \end{minipage}
      }
      \subfigure[I by ours]{
        \begin{minipage}[b]{0.124\linewidth}
        \includegraphics[width=\linewidth]{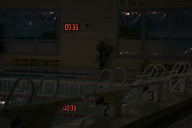}\vspace{6pt}
        \includegraphics[width=\linewidth]{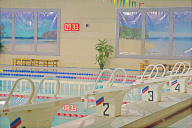}
        \end{minipage}
      }
      \caption{Decomposition results by LIME, Retinex-Net, and our method on the LOL dataset. The first column is low-light image, the second column is normal-light image.}
      \label{fig:decomposition}
    \end{figure*}

  In this section, we mainly evaluate the performance of our method on the LOw-Light~(LOL) dataset. The LoL paired dataset offered by Wei~\textit{et al.}~\cite{wei2018deep} contains 500 low and normal-light image pairs. Similar to the CSID dataset, the images of LOL are synthesized on the real scenes. The difference of two datasets is that the ground truth images of LOL are normal-light images while images in CSID are all captured on weak illumination. \\
  \textbf{Comparison of decomposition results.} In Fig.~\ref{fig:decomposition}, we compare the decomposition results by our method, the LIME method and the Retinex-Net method. It is worth noting that since the illumination image by our method is three-channel, the R and I results of decompression are the opposite of LIME and Retinex-Net. In fact, based on the analysis of Retinex theory, the results generated by our method are more in line with expectations because the amount of light received by each channel should be different. Aside from the influence of these factors, we can also find that the generated illumination images by us are more clearer than the reflected images by LIME and Retinex-Net. The estimated R images by LIME are different between low light and normal illumination while Retinex-Net try to reduce the differences of them but get a lot of noise when dealing with low-light image. By comparing with other methods, the Retinex-GAN seems to achieve better performance on decomposition when processing the same image.
 \begin{figure*}[htbp]
    \centering
    \subfigure[Input]{
      \begin{minipage}[b]{0.124\linewidth}
        \includegraphics[width=\linewidth]{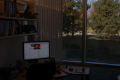}\vspace{6pt}
        \includegraphics[width=\linewidth]{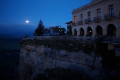}\vspace{6pt}
        \includegraphics[width=\linewidth]{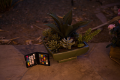}\vspace{6pt}
        \includegraphics[width=\linewidth]{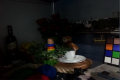}\vspace{6pt}
        \includegraphics[width=\linewidth]{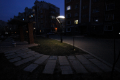}\vspace{6pt}
        \includegraphics[width=\linewidth]{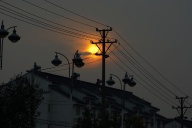}\vspace{6pt}
        \includegraphics[width=\linewidth]{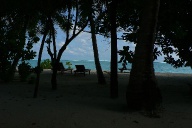}
       \end{minipage}

    }
    \subfigure[CLAHE]{
      \begin{minipage}[b]{0.124\linewidth}
        \includegraphics[width=\linewidth]{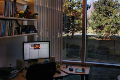}\vspace{6pt}
        \includegraphics[width=\linewidth]{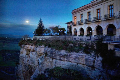}\vspace{6pt}
        \includegraphics[width=\linewidth]{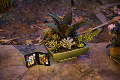}\vspace{6pt}
        \includegraphics[width=\linewidth]{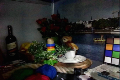}\vspace{6pt}
        \includegraphics[width=\linewidth]{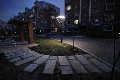}\vspace{6pt}
        \includegraphics[width=\linewidth]{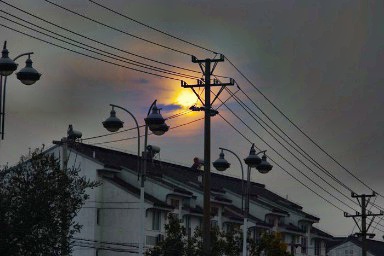}\vspace{6pt}
        \includegraphics[width=\linewidth]{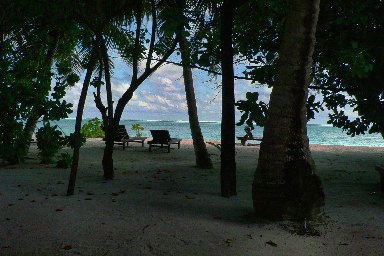}
     \end{minipage}

    }
    \subfigure[LightenNet]{
      \begin{minipage}[b]{0.124\linewidth}
        \includegraphics[width=\linewidth]{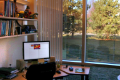}\vspace{6pt}
        \includegraphics[width=\linewidth]{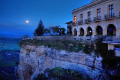}\vspace{6pt}
        \includegraphics[width=\linewidth]{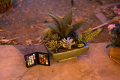}\vspace{6pt}
        \includegraphics[width=\linewidth]{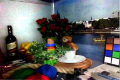}\vspace{6pt}
        \includegraphics[width=\linewidth]{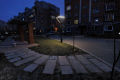}\vspace{6pt}
        \includegraphics[width=\linewidth]{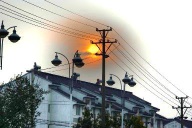}\vspace{6pt}
        \includegraphics[width=\linewidth]{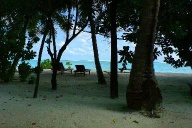}
      \end{minipage}
   }
    \subfigure[AWVM]{
      \begin{minipage}[b]{0.124\linewidth}
        \includegraphics[width=\linewidth]{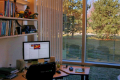}\vspace{6pt}
        \includegraphics[width=\linewidth]{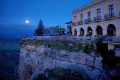}\vspace{6pt}
        \includegraphics[width=\linewidth]{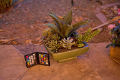}\vspace{6pt}
        \includegraphics[width=\linewidth]{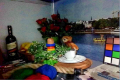}\vspace{6pt}
        \includegraphics[width=\linewidth]{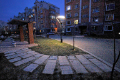}\vspace{6pt}
        \includegraphics[width=\linewidth]{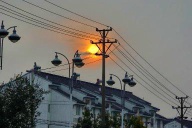}\vspace{6pt}
        \includegraphics[width=\linewidth]{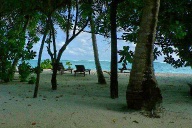}
         \end{minipage}
   }
    \subfigure[RetinexNet]{
      \begin{minipage}[b]{0.124\linewidth}
        \includegraphics[width=\linewidth]{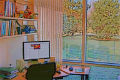}\vspace{6pt}
        \includegraphics[width=\linewidth]{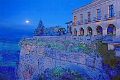}\vspace{6pt}
        \includegraphics[width=\linewidth]{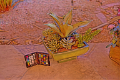}\vspace{6pt}
        \includegraphics[width=\linewidth]{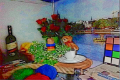}\vspace{6pt}
        \includegraphics[width=\linewidth]{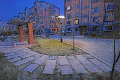}\vspace{6pt}
        \includegraphics[width=\linewidth]{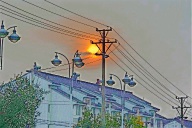}\vspace{6pt}
        \includegraphics[width=\linewidth]{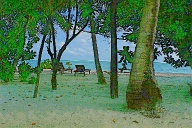}
     \end{minipage}
    }
	\subfigure[LIME]{
      \begin{minipage}[b]{0.124\linewidth}
        \includegraphics[width=\linewidth]{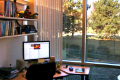}\vspace{6pt}
        \includegraphics[width=\linewidth]{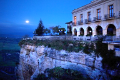}\vspace{6pt}
        \includegraphics[width=\linewidth]{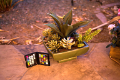}\vspace{6pt}
        \includegraphics[width=\linewidth]{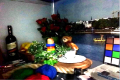}\vspace{6pt}
        \includegraphics[width=\linewidth]{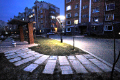}\vspace{6pt}
        \includegraphics[width=\linewidth]{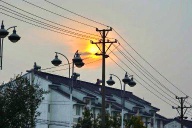}\vspace{6pt}
        \includegraphics[width=\linewidth]{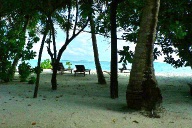}
      \end{minipage}
   }
    \subfigure[Ours]{
      \begin{minipage}[b]{0.124\linewidth}
        \includegraphics[width=\linewidth]{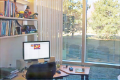}\vspace{6pt}
        \includegraphics[width=\linewidth]{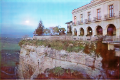}\vspace{6pt}
        \includegraphics[width=\linewidth]{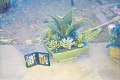}\vspace{6pt}
        \includegraphics[width=\linewidth]{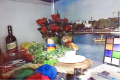}\vspace{6pt}
        \includegraphics[width=\linewidth]{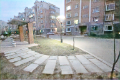}\vspace{6pt}
        \includegraphics[width=\linewidth]{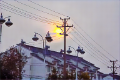}\vspace{6pt}
        \includegraphics[width=\linewidth]{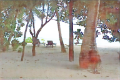}       
     \end{minipage}
   }
    \caption{Visual effects comparison on real scene dataset.}
    \label{fig:real}
\end{figure*}
\\\textbf{Comparison in the Real Scene.} According to above analysis, we hold that the illumination image enhanced by Retinex-GAN can be considered as a good enhancement result instead of the final synthesized result image. Therefore, we can also use the enhanced illumination image as the result of generation in practice. Then we evaluate the algorithm on the real scene while the evaluation dataset comes from public dataset. We compare Retinex-GAN with LIME, LightenNet, CLAHE, Retinex-Net and AWVM. The model of Retinex-GAN is trained on LOL dataset. In Fig.~\ref{fig:real}, it displays some of the experimental results and it is evident that the pictures tested were extremely dark. From the perspective of image brightness, the image generated by our method is much brighter than those ones by other methods. There are two main reasons. One is the data distribution. The brightness of low-light images in LOL dataset are generally low, and the ground truth images are very bright. This leads to the fact that the brightness of the generated image will increase the same level in the brightness level of the tested image when processing actual image. Another reason is that, as we mentioned earlier, we take the brightness of the intermediate enhancement as the final result while the enhanced illumination image is a little brighter than the final composite image inherently. Although the increase in brightness leads to the weakening of color, the results we produced are visually acceptable, and more comfortable than those generated by other methods in some ways. 

\subsection{Ablation Study on CSID}
\begin{figure}[htbp]
    \centering
       \subfigure[]{
        \begin{minipage}[b]{0.466\linewidth}
        \includegraphics[width=\linewidth]{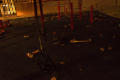}
        \end{minipage}
      }
      \subfigure[]{
        \begin{minipage}[b]{0.466\linewidth}
        \includegraphics[width=\linewidth]{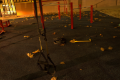}
        \end{minipage}
      }
        \begin{minipage}[b]{0.03\linewidth}
        S1\vspace{27pt}
        
        S2\vspace{26pt}
        
        S3\vspace{14pt}
        \end{minipage}
      \subfigure[]{
        \begin{minipage}[b]{0.2\linewidth}
        \includegraphics[width=\linewidth]{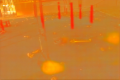}\vspace{6pt}
        \includegraphics[width=\linewidth]{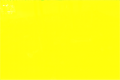}\vspace{6pt}
        \includegraphics[width=\linewidth]{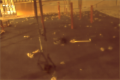}
        \end{minipage}

      }
      \subfigure[]{
        \begin{minipage}[b]{0.2\linewidth}
        \includegraphics[width=\linewidth]{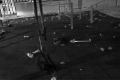}\vspace{6pt}
        \includegraphics[width=\linewidth]{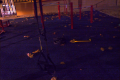}\vspace{6pt}
        \includegraphics[width=\linewidth]{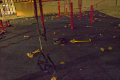}
        \end{minipage}

      }
      \subfigure[]{
        \begin{minipage}[b]{0.2\linewidth}
        \includegraphics[width=\linewidth]{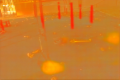}\vspace{6pt}
        \includegraphics[width=\linewidth]{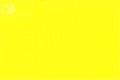}\vspace{6pt}
        \includegraphics[width=\linewidth]{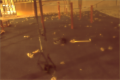}
        \end{minipage}

      }
      \subfigure[]{
        \begin{minipage}[b]{0.2\linewidth}
        \includegraphics[width=\linewidth]{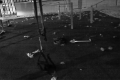}\vspace{6pt}
        \includegraphics[width=\linewidth]{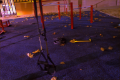}\vspace{6pt}
        \includegraphics[width=\linewidth]{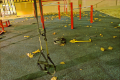}
        \end{minipage}
      }  
    \caption{Decomposition results by S1, S2 and S3 in the CSID datasets: (a)~Input images. (b)~Ground truth. (c)~Reflected image of (a). (d)~Illumination image of (a). (e)~Reflected image of (b) (f)~Illumination image of (b). From top to bottom: results by S1, S2 and S3.}
    \label{fig:S2S3}
  \end{figure}

  The scientific contribution of our work could come in three aspects. Firstly, so far we're the first to combine the Retinex theory with GAN for the research of low-light image enhancement as we know. Secondly, we propose to add a regularization loss function for the decomposition loss adapted to Retinex theory. Moreover, some of the latest tricks were employed to make the network stable. In order to show the influence of each of these contributions, we conduct the following experiments. First, we build three basic networks which implement three different strategies on chapter~\ref{ref:regular} without GAN, SSIM-loss and smooth-L1 loss. Then we add new components one by one and observe the changes in evaluation values on the basis of the third strategy. We did ablation experiments on CSID with a brightness level of 0.5. In Tab.~\ref{tab:ablation}, S1, S2 and S3 represents the first, second and third strategy respectively. As can be seen from the table, since the decomposition solution space of S2 contains the decomposition solution space of S1, the values of PSNR and SSIM have increased while the values of MSE have also declined by 6.2. Then, in Fig.~\ref{fig:S2S3}, the more meaningful reflected and illumination images generated by S3 fully proves that the network with regularization loss function are more consistent with Retinex theory than S1 and S2. Simultaneously, adding GAN loss, SSIM loss and using Smooth-L1 loss instead of normal L1 loss also improves network performance.
\begin{table}
\renewcommand\arraystretch{1.9}
	\centering	
	\caption{}
	\begin{tabular}{|p{0.45\linewidth}|c|c|c|}
  		\hline  	
  		&PSNR & MSE & SSIM\\\hline
  		S1 &30.54&111.27&0.853\\\hline
  		S2 &30.76&105.07&0.859\\\hline
  		S3 &30.89&103.03&0.864\\\hline
  		S3 + Smooth-L1 + SSIM&30.87&103.41&0.872\\\hline
  		S3 + Smooth-L1 + SSIM + GAN&\textbf{31.31}&\textbf{99.12}&\textbf{0.879}\\\hline    
  	\end{tabular}  
  	\label{tab:ablation}  
\end{table}

\subsection{Discussion}

\begin{figure}[htbp]
    \centering
       \subfigure[]{
        \begin{minipage}[b]{0.21\linewidth}
        \includegraphics[width=\linewidth]{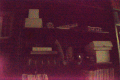}
        \end{minipage}

      }
      \subfigure[]{
        \begin{minipage}[b]{0.21\linewidth}
        \includegraphics[width=\linewidth]{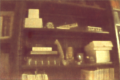}
        \end{minipage}

      }
      \subfigure[]{
        \begin{minipage}[b]{0.21\linewidth}
        \includegraphics[width=\linewidth]{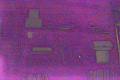}
        \end{minipage}

      }
      \subfigure[]{
        \begin{minipage}[b]{0.21\linewidth}
        \includegraphics[width=\linewidth]{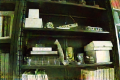}
        \end{minipage}

      }
      \subfigure[]{
        \begin{minipage}[b]{0.21\linewidth}
        \includegraphics[width=\linewidth]{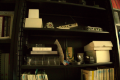}
        \end{minipage}

      }
      \subfigure[]{
        \begin{minipage}[b]{0.21\linewidth}
        \includegraphics[width=\linewidth]{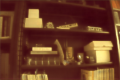}
        \end{minipage}
      }
      \subfigure[]{
        \begin{minipage}[b]{0.21\linewidth}
        \includegraphics[width=\linewidth]{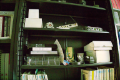}
        \end{minipage}
      }
       \subfigure[]{
        \begin{minipage}[b]{0.21\linewidth}
        \includegraphics[width=\linewidth]{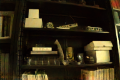}
        \end{minipage}
      }
    \caption{Failed examples in the CSID datasets: (a)~Input images with a lot of noise. (b)~Reflected image of (a). (c)~Illumination image of (a). (d)~Enhanced illumination image of (a). (e)~Ground truth. (f)~Reflected image of (e). (g)~Illumination image of (e). (h)~Enhanced image of (a).}
    \label{fig:failures}
  \end{figure}

Strong though the proposed hybrid Retinex-GAN is, there remains still much work to be done to develop our ideas. Firstly, maybe we will be likely faced with the shortage of paired inputs including low and high images which are not so easy to acquire in the real environment. A small scale of datasets is unable to maximize the performance of such a deep network. As such, combining the Retinex theory with unsupervised learning algorithm will be considered in future work. Secondly, the established complex GANs with Retinex theory in this work operated at the speed of 91 frames per second when dealing with image of ${384}\times{256}$ resolution on GTX1080TI but only got 11 FPS at the resolution of ${1280}\times{720}$,  so that the processing efficiency will not suffice to cope with the real-time video processing in reality.  So, the optimization of the networks will be worth studying in the next step. Furthermore, even the most advanced algorithm will become useless when dealing with the extremely weak light images and the ambient noise is too high enough. Hence, more preparation in the image pre-processing stage should be guaranteed to prevent the invalid dataset. Here the authors present some of the failed examples in~Fig.\ref{fig:failures}. Since the input image (a) contaminated by much noise, the decomposition images (b) and (c) by our techniques were consequently full of noise. Although the model is possible to restore images (h) similar to the ground truth (e), there’s still much missing details such as words on the book cover. In brief, the authors had a great confidence that such a novel work will be of considerable value in image processing application.

\section{Conclusion}
By combining generative adversarial network with Retinex theory, low-light image enhancement issue was settled down effectively. To improve the quality of output images, the authors had tried to introduce Structural Similarity loss to avoid the side effect of blur and to provide a global optimization possibility. The convincing tests and satisfactory results have encouraged the authors to carry on the further study to investigate the application of Retinex-GAN to the scenario of higher resolution video streams and images. We believe in that such a hybrid architecture of GAN and Retinex Theory will undoubtedly benefit the image processing.

%

 \ifCLASSOPTIONcompsoc
   \section*{Acknowledgments}
 \else
   \section*{Acknowledgment}
 \fi

The authors will be grateful to the referees and the Editor for their valuable time and contributions. This work is supported partly by the Anhui Provincial Natural Science Foundation under Grant 1908085QF254 and the Fundamental Research Funds for the Central Universities under Grant WK2100060023. Furthermore, we wish to acknowledge the faculty of Key Laboratory of Network Communication System and Control who made this research possible. We mostly appreciated their helpful suggestions.

 \ifCLASSOPTIONcaptionsoff
   \newpage
 \fi


\bibliographystyle{IEEEtran}
\bibliography{IEEEabrv,mybib}


%
%
%

\end{document}